\documentclass[10pt,twocolumn,letterpaper]{article}

\usepackage{cvpr}              

\usepackage{graphicx}
\usepackage{amsmath}
\usepackage{amssymb}
\usepackage{booktabs}
\usepackage{multirow}

\usepackage[pagebackref,breaklinks,colorlinks]{hyperref}

\usepackage[capitalize]{cleveref}
\crefname{section}{Sec.}{Secs.}
\Crefname{section}{Section}{Sections}
\Crefname{table}{Table}{Tables}
\crefname{table}{Tab.}{Tabs.}

\def\cvprPaperID{21} 
\def\confName{CVPR}
\def\confYear{2023}

\begin{document}

\title{Deep Prototypical-Parts Ease Morphological Kidney Stone Identification and are Competitively Robust to Photometric Perturbations}

\author{Daniel Flores-Araiza$^{1}$
\and
Francisco Lopez-Tiro$^{1}$
\and
Jonathan El-Beze$^{2}$
\and
Jacques Hubert$^{2}$
\and
Miguel Gonzalez-Mendoza$^{1}$
\and
Gilberto Ochoa-Ruiz$^{1}$
\and
Christian Daul$^{3}$
\and \\
%
%
%
$^{1}$Tecnologico de Monterrey, School of Engineering and Sciences, Mexico\\
$^{2}$CHU Nancy, Service d’urologie de Brabois, Nancy, France\\
$^{3}$CRAN UMR 7039, Université de Lorraine and CNRS, Nancy, France\\ \\
}

\maketitle

\begin{abstract}
Identifying the type of kidney stones can allow urologists to determine their cause of formation, 
improving the prescription of appropriate treatments to diminish future relapses. 
Currently, the associated ex-vivo diagnosis (known as Morpho-constitutional Analysis, MCA)
 is time-consuming, expensive and requires a great deal of experience, as it requires a visual analysis component that is highly operator dependant. 
Recently, machine learning methods have been developed for in-vivo endoscopic stone recognition. 
Deep Learning (DL) based methods outperform non-DL methods in terms of accuracy but lack explainability. 
 Despite this trade-off, when it comes to making high-stakes decisions, it's important to prioritize understandable Computer-Aided Diagnosis (CADx) that suggests a course of action based on reasonable evidence, rather than a model prescribing a course of action.
%
%
In this proposal, we learn Prototypical Parts (PPs) per kidney stone subtype, which are used by the DL model to generate an output classification. Using PPs in the classification task enables case-based reasoning explanations for such output, thus making the model interpretable. In addition, we modify global visual characteristics to describe their relevance to the PPs and the sensitivity of our model's performance. With this, we provide explanations with additional information at the sample, class and model levels in contrast to previous works. 
Although our implementation's average accuracy is lower than state-of-the-art (SOTA) non-interpretable DL models by 1.5\%, our models perform 2.8\% better on perturbed images with a lower standard deviation, without adversarial training. Thus, Learning PPs has the potential to create more robust DL models. \textit{Code at:} \url{https://github.com/DanielF29/Prototipical_Parts}


\end{abstract}


\begin{figure*}[ht] 
    \centering
    \includegraphics[width=0.90\linewidth]{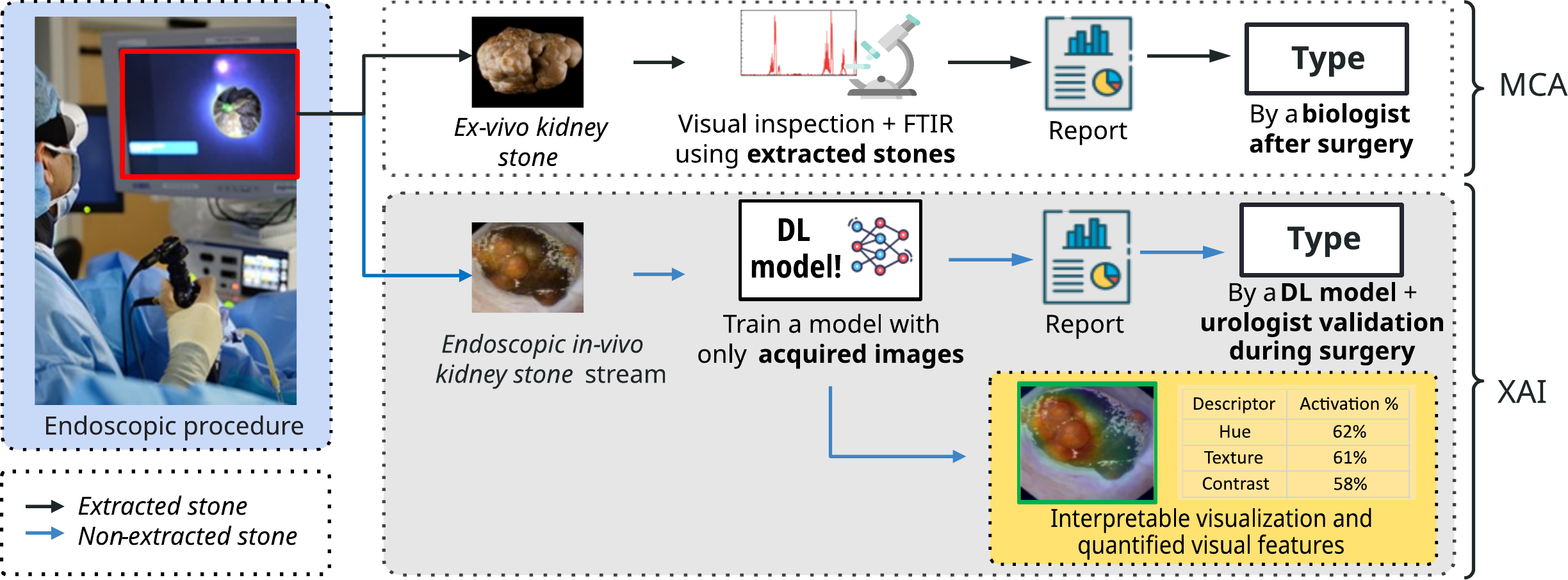} 
    \caption{\small{
    Traditional Morpho-constitutional Analysis (MCA) involves an endoscopic procedure to locate and extract a kidney stone (Ex-vivo), and subject it to visual and Fourier Transform Infrared Spectroscopy (FTIR) analysis by a specialist for classification, in the MCA section of the figure.
    Recent works have proposed using Deep Learning (DL) to classify kidney stones during an endoscopic procedure (In-vivo) automatically, represented in the XAI section of the figure.
    We argue and show, eXplainable AI (XAI) is needed for the adoption of DL as a Computer Assisted Diagnosis (CADx) tool. In the orange box is the extension to the current DL systems this proposal enables.
    }}
    \label{fig_MCA}
\end{figure*}

\begin{figure*}[ht] 
    \centering
    \subfloat[\small{Model architecture and proposed label}]{
    \label{fig_MI_XAI_PPs_diagram_a}
    \includegraphics[width=0.44\linewidth]{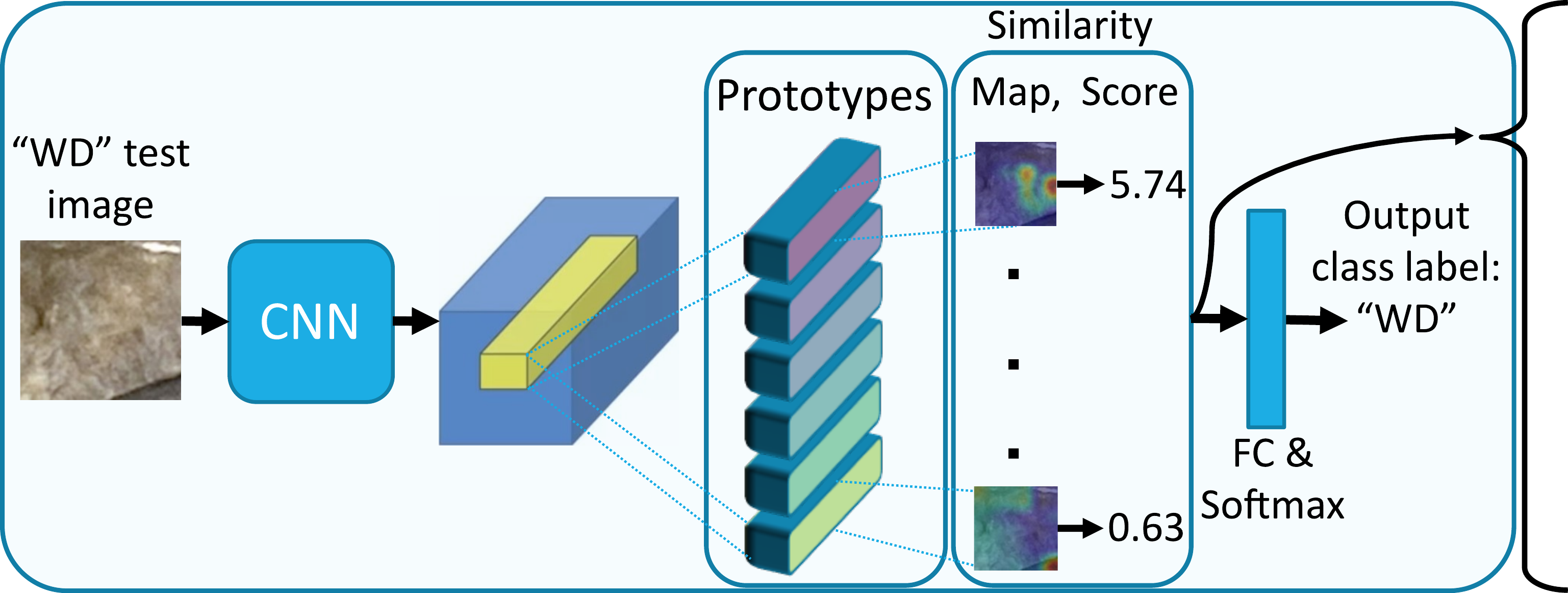} 
    } 
    \subfloat[\small{Visual explanations, similarity scores and visual quantitative descriptors}]{
    \label{fig_MI_XAI_PPs_diagram_b}
    \includegraphics[width=0.55\linewidth]{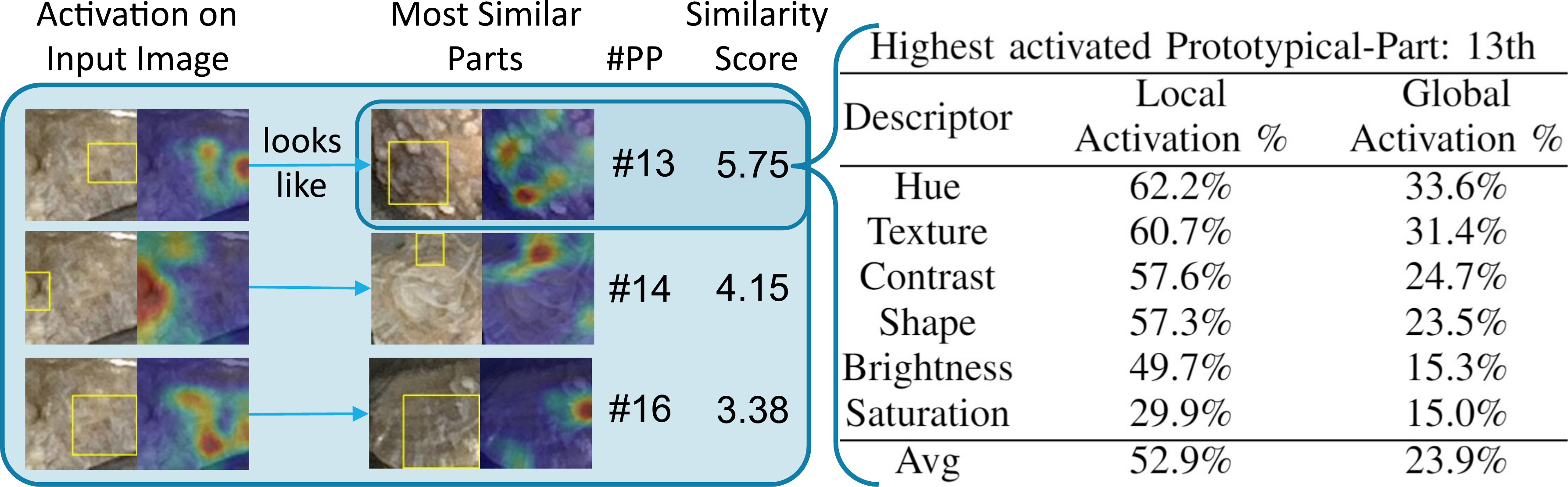} 
    }
    \caption{\small{
    (a) The model classifies correctly a kidney stone patch, sub-type "WD", by measuring the similarity of learned Prototypical-Parts (PPs) to parts in the input image.
    As seen in (b), 
    first, the most relevant area on the input image is indicated by a yellow bounding box and a heatmap representing levels of similarity on the input image for each Prototypical-Part (PP).
    Second, the image with the representative part from the training set that \textit{looks like} the input indicated part is also shown on a yellow box, with its corresponding heatmap.
    Third, the level of saliency of visual characteristics and the descriptors are obtained. 
    In this example descriptors for the most relevant PP are shown, indicating hue as the most relevant feature for the activation of this PP. 
    Then, PPs identified work both as our explanations and for the classification, rendering this architecture interpretable, easing the urologist task of kidney stone classification, as they are provided with the reasons used by the model for the suggested classification. 
    }}
    \label{fig_MI_XAI_PPs_diagram}
\end{figure*}
    
\section{Introduction}
\label{introduction}


Urolithiasis is a disease in which kidney stones form somewhere in the urinary tract \cite{viljoen2019renal}.
In developed countries, this disease has become a public health problem and presents a high incidence of kidney stones episodes, affecting up to 10\% of their population \cite{nassir2018impact}.
The formation of kidney stones is strongly related to factors such as diet \cite{friedlander2015diet, siener2003fluid}. However, there are other factors that can lead to the production of a kidney stone such as age, chronic diseases, hereditary-family history, among others \cite{corrales2021classification, daudon2018recurrence}.
The timely identification of a kidney stone aids clinical specialists in determining the causes of its formation and to prescribe a personalized treatment, which can prevent relapses \cite{friedlander2015diet}. 

The morpho-constitutional analysis (MCA) is the standard technique for determining the different types of whole kidney stones removed during surgery (ex-vivo stones) \cite{daudon2016comprehensive, daudon2018recurrence}.  
Using this method it is possible to identify 21 different types of kidney stones, with both pure and mixed compositions \cite{corrales2021classification}. 
Although the MCA efficiently establishes the type of ex-vivo kidney stones, it is very difficult to provide a reliable diagnosis during an endoscopic intervention (in-vivo), and the results of the MCA on extracted stones (ex-vivo) may take several days. 
Also, it can be difficult to carry out the inspection during surgery, as the extraction process 
can take up to one hour \cite{corrales2021classification}. Furthermore, it is very difficult to train specialists on this technique, even considering a high incidence of kidney stone episodes. 
%

%


Given the importance and difficulty of performing the visual inspection in a repeatable and highly reliable manner, 
recently, several Artificial Intelligence (AI) methods have been proposed for automating the kidney stone classification process.
Using AI to automate the classification of kidney stones entails either replacing MCA or aiding the surgeon during the extraction process.\\

Several DL-based techniques in particular have demonstrated that is possible to replicate the results obtained by well-trained specialists \cite{lopez2021assessing, estrade2021towards, black2020deep, ochoa2022vivo, Onal2022AssessingKS}.
%
However, DL models lack interpretation of the extracted features, making these models of limited help in clinical settings.
To facilitate clinical acceptance, 
a DL model will need to provide the specialist with its reasoning process, this implies providing evidence on the \textit{what}, \textit{how} and \textit{why} behind any suggested classification \cite{FDA_edwards_2019, CNNs_for_radiologist_A_guide_2019}. 
Therefore, to be a useful aid for CADx purposes, any evidence should reflect the same key distinctions used for MCA by a specialist. 

In order to pave the way for AI-based interpretable MCA using deep learning techniques, we leverage recent strides in explainability that seek to base image classification on case-based reasoning \cite{ProtoPnet, ProtoPshare, AIAIBL}. 
In this work, both visual explanations and quantitative information about visual characteristics deemed important by the network are provided \cite{PP_Descriptors}.
It must be emphasized that the explanations provided by our approach follow the reasoning processes of urologists during MCA by showing the detected morphological relevant features of each image as learned PPs. 

Overall, this work presents an interpretable DL classifier for MCA.
This proposal facilitates human-machine collaboration for morphological analysis whose explicit reasoning can be easily inspected, understood and verified by urologists and medical specialists. In contrast to previous works, we evaluate our interpretable models and their DL counterparts under perturbations of global visual characteristics to describe their relevance to the PPs and the sensitivity of our model.

The of the paper is organized as follows. In the following Section \ref{Proposed_Approach} the proposed approach reasoning and justification are presented. In Section \ref{Experimental_Setup} the experimental setup is described for our experiments. In Section \ref{Results} we share our results and present a discussion on their implications, sharing one explanation case comparing a correct and incorrect classification. Finally, in Section \ref{Conclusions} the conclusion and the insights obtained for further research in the field are presented.
\section{Motivation}\label{Proposed_Approach}
\begin{figure*}[ht] 
    \centering
    \includegraphics[width=1.0\linewidth]{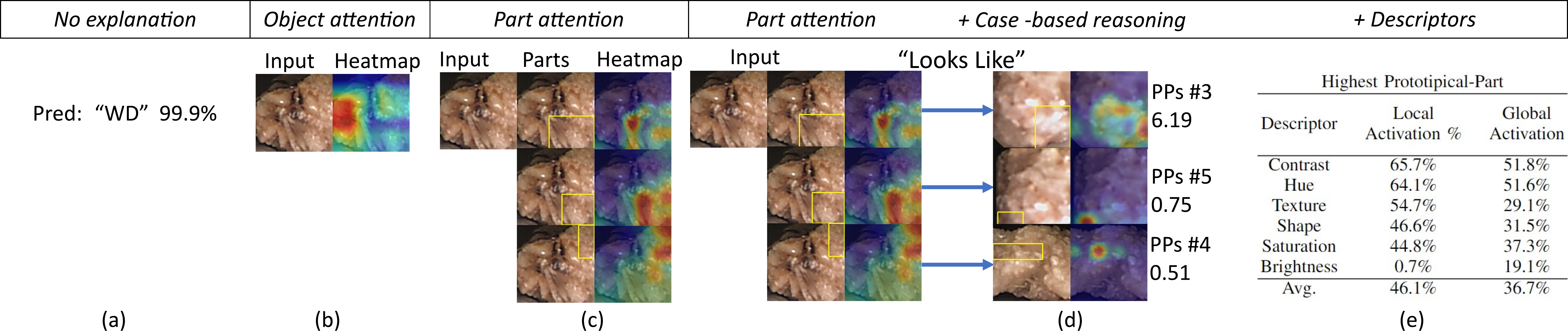}
    \caption{\small{
    Visual comparison of different types of model interpretability on an example of a kidney stone image as an example. (a) Traditional non-interpretable deep learning output classification of a kidney stone \cite{lopez2021assessing},  
    (b) object-level attention map (e.g., class activation map \cite{GradCAM}, (c) part attention (provided by attention-based interpretable models \cite{Fine_Grained_Zheng_2017_ICCV}); and (d) part attention with similar prototypical parts \cite{ProtoPnet}, with additional (e) quantitative evaluations of visual features, the \textit{descriptors} \cite{flores_araiza2022interpretable_KS, PP_Descriptors}.
    }}
    \label{XAI_types}
\end{figure*}
MCA consists of two complementary analyses of the kidney stone ex-vivo in the laboratory. 
First, a specialist performs a visual inspection of the characteristics (such as color, texture and shapes) of the surface (external) and section (cross-sectional) views. 
Second, a biochemical study is performed to determine the main components in the stone through an Fourier Transform Infrared Spectroscopy (FTIR) study \cite{estepa1997contribution}.
Both studies complement each other to provide a detailed report to determine the type of kidney stone.
This traditional MCA process is represented in the MCA area in Fig. \ref{fig_MCA}.

Recently proposed AI systems for kidney stone classification are presented as automated CADx tools. This approach could enable faster and more reliable diagnosis based solely on kidney stone images, reducing the need for invasive procedures, on in-vivo \cite{estrade2021towards} or ex-vivo \cite{lopez2021assessing, black2020deep, ochoa2022vivo} conditions, as represented in \textit{MCA-Based DL}, the grey area at Fig. \ref{fig_MCA}. 
Also, for in-vivo scenarios, the operations would be quicker to perform and less traumatic, due to the classification of the kidney stones right before performing dusting, a process intended to fragment and destroy the kidney stones inside the urinary tract. 
%
Current DL classifiers of kidney stones are black boxes, as those have the limitation of being unable to describe the inner workings that led to a given prediction beyond the class label.
Therefore, these types of models cannot provide useful information to the specialist to understand how features on the input image were used to output the classification proposed by the DL model. 

The eXplainable AI (XAI) field seeks to remedy these limitations.
XAI methods are used to provide an explanation of how DL models arrived at a particular output, typically after the classification was assigned. 
If DL models achieve high performance and can provide explanations for the classification based on relevant features for the specialists, then it would be easier to maintain accountability for recommended treatments based on those analyses. 
In the area indicated by the XAI bracket at Fig. \ref{fig_MCA}, a holistic explanation is represented, by providing the description of what and why visual features are relevant from the input image, as quantitative evaluations of visual features, and a saliency map highlighting where the described features are located. 
Explanations that could be considered complete do not yet exist for kidney stone identification. Nonetheless, improving explanations by indicating precise visual characteristics would facilitate the adoption of DL CADx tools, resulting in faster, more accountable, and repeatable diagnosis recommendations for the treatment of urolithiasis.

Taking into account the previously mentioned desired characteristics of explanations, in this work, we seek to improve the current methods by adding the required explainability to facilitate the morphological analysis of kidney stones, our proposal of how to achieve this is depicted in Fig. \ref{fig_MI_XAI_PPs_diagram}.
For context, Fig. \ref{XAI_types} presents a visual comparison of different explanations.
Then, to improve DL-generated explanations for the classification of kidney stones we began generating saliency maps of learned representative parts, the learned PPs, for each class \cite{flores_araiza2022interpretable_KS}, after, we evaluate which global visual characteristics PPs are relaying for such classification \cite{PP_Descriptors}.

\section{Proposed Approach}\label{ProtoPNet_plus_descriptors}
Having a single heatmap as a visual explanation, as most current XAI visualization methods do, is an oversimplification of all the characteristics expected and used to classify a kidney stone into its corresponding type.
To alleviate the unmet needs of most of the current DL models and XAI visualization methods, this proposal leverages a case-based reasoning process.
Our proposal pipeline does this by extracting semantic features, representing parts, from an input image with a Convolutional Neural Network (CNN).
Then, the extracted features are measured against the learned prototypical parts, in the form of similarity scores, as indicated by equation \ref{eq_similarity_score}, and represented in Fig. \ref{fig_MI_XAI_PPs_diagram_b}.
Finally, the output prediction of the classification task is a weighted combination of the similarity scores.
Using these same learned PPs for the output of the model and the explanations, the method guarantees faithfulness of the explanations to the inner workings of the model.
Current case-based reasoning methods, instead of using high-dimensional features from the convolutional layers of a CNN, use the identification of a few prototypical parts \cite{ProtoPnet, AIAIBL, ProtoPshare, 2021Differentiable_ProtoPNet}.
This identification of a limited number of prototypical parts enables user understanding of the learned reasoning behind the DL model output, which we show, achieves competitive performance against its non-interpretable counterparts. 
The advantage of the case-based reasoning process is that the explanations generated follow the same reasoning process a specialist used to classify a kidney stone image.
However, for images of kidney stones, a great level of expertise is required, and thus, explanations generated from PPs for kidney stone patches still depend on non-obvious characteristics from the input image for non-experts, reason for us to quantify the sensitivity of PPs to a set of perturbations  \cite{PP_Descriptors}, which we dub herein as ``Descriptors".

Each perturbation is directed at a particular visual characteristic (Brightness, Saturation, Shape, Texture, Contrast, and Hue). 
These descriptors allow measuring individually the level of importance of visual features to each of the PPs, as shown in the table in Fig. \ref{fig_MI_XAI_PPs_diagram_b}.

%
Our model shows the underlying decision-making process for each case, based on several explanations which correspond to the most similar cases identified and used by the model to give its suggested classification of the image.
Fig. \ref{fig_MI_XAI_PPs_diagram} shows an example of how our proposal works: 
\textbf{I)} Our model processes an input image through a feature extractor backbone (CNN in Fig. \ref{fig_MI_XAI_PPs_diagram_a}).
\textbf{II)} Similarity of the learned PPs to the extracted features from the CNN are calculated accordingly to equation \ref{eq_similarity_score} (Similarity score in Fig. \ref{fig_MI_XAI_PPs_diagram_a}).
\textbf{III)} From the Prototypes layer, in Fig. \ref{fig_MI_XAI_PPs_diagram_a}, the scores of PPs calculated against the input image are processed through a linear weighted combination, a fully connected layer, to provide a final classification of a kidney stone image.
\textbf{IV)} The main PPs that determined the classification are provided as the most similar cases to their respective parts from the input image, as seen in the blue box in Fig. \ref{fig_MI_XAI_PPs_diagram_b}.
\textbf{V)} Finally, the saliency of certain characteristics for each prototypical part is evaluated, the \textit{Descriptors}, listed in the table in Fig. \ref{fig_MI_XAI_PPs_diagram_b} and provide a description of the input image features that mainly triggered the activation of the corresponding learned Prototypical-Part (PP).

This capacity yields predictions that are easy to understand by specialists familiarized with the features related to each kidney stone type and subtype, rendering classifications interpretable for urologists. 
Thus, aiming to help in the morphological analysis of each kidney stone image, the trained model provides the specialist with the means to trust the AI, check its output for plausibility, and easily overrule it when necessary.
Our approach, with its inherently interpretable reasoning process, contrasts directly with previous work that relied on post-hoc explanation techniques to explain a trained black-box model on particular classifications \cite{ochoa2022vivo, Towards_autoKSclass_Vincent2021} or with global explanations \cite{elbeze2022}.

%
Following the architecture proposed on \cite{ProtoPnet, ProtoPshare}, 
in this study, we explore ProtoPNet architecture with some important differences, 
\textbf{I)} the analysis performed explores different backbones, with the intention to compare CNN with different levels of intricate connections. 
\textbf{II)} Different number of prototypical parts per class and 
\textbf{III)} The relevance of data augmentation for training.
A relevant note is that the training of the models was performed without part annotations, only relying on the class label.
\begin{table}[]
\centering
\caption{Number of complete endoscopic images acquired.}
\vspace{-0.15cm}
\label{tab:dataset}
\resizebox{\columnwidth}{!}{%
\begin{tabular}{@{}cccccc@{}}
\toprule
Subtype & Main component & Key & Surface & Section & Mixed \\ \midrule  \vspace{-0.05cm}
Ia & Whewellite & WW & 62 & 25 & 87 \\  \vspace{-0.05cm}
IIa & Weddellite & WD & 13 & 12 & 25 \\  \vspace{-0.05cm}
IIIa & Acide Urique & AU & 58 & 50 & 108 \\  \vspace{-0.05cm}
IVc & Struvite & STR & 43 & 24 & 67 \\  \vspace{-0.05cm}
IVd & Brushite & BRU & 23 & 4 & 27 \\  \vspace{-0.05cm}
Va & Cystine & CYS & 47 & 48 & 95 \\ \cmidrule(l){3-6}   \vspace{-0.05cm}
 &  & TOTAL & 246 & 163 & 409 \\ \bottomrule  \vspace{-0.05cm}
\end{tabular}
}
\end{table}

\begin{figure}[]
\centering
    \includegraphics[width=0.90\linewidth]{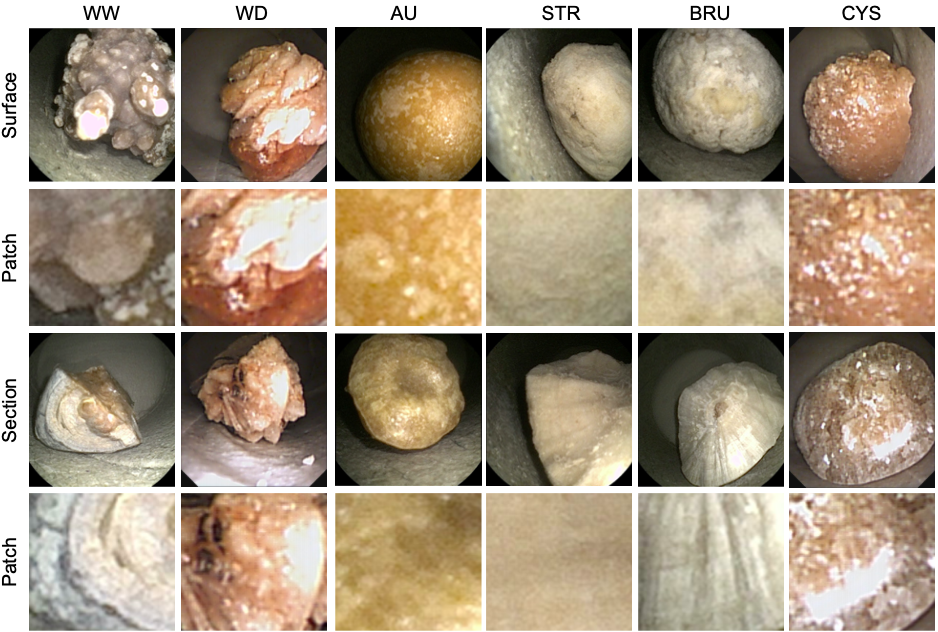} 
    \caption{\small{Examples of kidney stones patches extracted from the dataset used in \cite{elbeze2022}. This endoscopic dataset consists of six of the most common kidney stones namely Whewellite (WW), Weddellite (WD), Acide Urique (AU),  Struvite (STR),  Brushite (BRU), and Cystine (CYS).}} 
\label{fig:im1}
\end{figure}
%
\subsection{Kidney stone dataset}\label{Kidney_stone_dataset}
%
A simulated in-vivo dataset of endoscopic images was used for the experiments reported in this contribution (Table \ref{tab:dataset}). This dataset consists of surface (246 images) and section (163 images) views. Each view contains six of the most common kidney stone sub-types: Ia (Whewellite, WW), IIa (Weddellite, WD), IIIa (Uric Acid anhydrous, UA), IVc (Struvite, STR), IVd (Brushite, BRU), and Va (Cystine, CYS).
Two different reusable digital flexible ureteroscopes Karl Storz (Storz Image 1 Hub and Storz image1 S) captured images of real kidney stones in an environment that simulates real conditions (such as illumination or blur) during surgery. In addition, the acquired endoscopic images were labeled manually by the urologist Jonathan El-Beze (see \cite{elbeze2022} for more details).

In order to train DL-based models, 12,000 square-patches of 256$\times$256 pixels were extracted from 409 endoscopic images of both views. This combined data is referred to as \textit{mixed} dataset. Patches of the dataset are shown in Fig. \ref{fig:im1}. 
For more details, patches were extracted with the method described in \cite{ochoa2022vivo} and \cite{lopez2021assessing}.
%

\subsection{Model architecture:}\label{Model_architecture:}
The architecture implemented uses a feature extractor $f(x)$, in this proposal a CNN, to perform the extraction of semantic features from the input images $x$ and learn $n$ prototypes per class, corresponding to parts in the input image. 
This first step of feature extraction from the input image implies the potential diversity and quality of possible prototypical parts to be learned from.
Hence, three CNN backbones are explored (VGG16, ResNet50 and DenseNet201) in order to compare the different possible performances obtained with our approach and compared against pure non-interpretable CNNs present in the estate of the art for kidney stone identification. 

Next, two layers of $1\times1$ convolutions are added, to adjust the depth of the feature activation maps to the depth selected of 128 for the layer of PPs, layer $g$.
The variable $l$ is used to index each of the $128 \times 1 \times 1$ patches $f(x)_l$ across the spatial dimensions.
The layer of PPs $g$, contains $n$ prototypes to be learned per class, with dimensions $128 \times 1 \times 1$.

Since a prototype has the same number of channels but a smaller spatial dimension
than the convolutional feature maps $f(x)$, we can interpret the prototype as 
representing a prototypical activation pattern of its class and we can visualize 
the prototype as a patch of the training image it appears in.
The distance between $p_j$ and each of the $l$ $1 \times 1$ spatial patches of the convolutional feature
map $f(x)$ is measured by $d_{j,l} = \lVert  p_j - f(x)_{l} \rVert_{2}^{2}$, and converts these distances to similarities $s_{j,l}$ by:   
\begin{equation}\label{eq_similarity_score}
    s_{j,l}= log \frac{d_{j,l}+1}{d_{j,l}+\epsilon}
\end{equation}
Where $\epsilon$ is a small value to avoid division by zero. This similarity scores $s_{j,l}$ can be arranged spatially into a similarity map, upsampled and superimposed on the input image to generate a saliency map as the ones presented on the blue box in Fig. \ref{fig_MI_XAI_PPs_diagram_b}.
The overall similarity score of the corresponding PPs is the highest score $s_{j}$ obtained from $s_{j,l}$.
These similarity scores are then processed with a fully connected layer, as to combine them in a weighted combination to obtain an output per class, which is then normalized with a softmax function. The details for the initialization and training hyperparameters, except for the data augmentation used, follow \cite{ProtoPnet}.

\subsection{Descriptors calculation:}\label{Descriptors_calculation}
Importance scores for image characteristics of the PPs are analyzed by measurement of the change in the similarity score $s_{j}$ when perturbing the input image with certain modifications $\hat{s_{j}}$.
Let $i\in\left\{brightness, contrast, saturation, hue, shape, texture\right\}$  denote the type of modification, as further detailed in \cite{PP_Descriptors}. 
Then the local importance score $\Phi_{local}^{i,j,k}$ of characteristic $i$ for test image 
$j \in S_{test}$ with class $k$ on the $j-th$ prototype is the difference in similarity score:
\begin{equation}\label{Local_descriptors}
    \Phi_{local}^{i,j,k} = s_{j} - \hat{s_{j}}
\end{equation}
And its global importance is a weighted arithmetic mean by weighing the local scores of all images in $S_{train}$ by their respective original similarity score per PP:
\begin{equation}\label{Global_descriptors}
    \Phi_{global}^{i,j} = \frac{\sum_{k=1}^{|S_{train}|} \Phi_{local}^{i,j,k} \cdot s_{j} }{\sum_{k=1}^{|S_{train}|} s_{j}} 
\end{equation}
Hence, if unmodified image $k$ gets a low similarity score with prototype $j$, it will provide a low weight during the global importance calculation.
In contrast, if the prototype $j$ is clearly present in image $k$, the architecture will assign a high similarity score and hence the local importance score for image $k$ gets a high weight.
As demonstrated in equation \ref{Global_descriptors}, these importance scores can be used to create global explanations that explain a prototype, and local explanations, equation \ref{Local_descriptors}, that explain
the main visual features responsible for a high similarity score between a given image and a prototype.
As appreciated from equation \ref{Local_descriptors} and equation \ref{Global_descriptors}, with this \textit{descriptors} the intention is to explain the identification of PPs on the input image from the models perspective, not from human perception, nonetheless, since the \textit{descriptors} explored are humanly interpretable actually those do enrich the explanations obtained.
The specifics for the different perturbations used to obtain the values of the PPs \textit{descriptors} follow \cite{PP_Descriptors}.
\subsection{Evaluation metrics}\label{Evaluation_metrics}
%
To determine the different types of kidney stones in endoscopic images, MCA specialists perform a visual inspection of the surface (external part) and  section (internal part) view, and provide a detailed report of each view of the kidney stone. 
In addition, in order to improve performance in this task, MCA also proposes combining information from surface and section views, since both views provide relevant information for classification. \cite{corrales2021classification} 

In this work, we perform different training of mixed views  models (similar to how MCA specialists perform it).
The accuracy and precision metrics are mainly used to quantitatively evaluate different implementations of DL models proposed in the SOTA mentioned to be intended as a CADx tool \cite{black2020deep, estrade2021towards, lopez2021assessing, ochoa2022vivo, torrell2018metric}. Therefore, our particular implementations were measured and compared on accuracy as well.
It can be observed in Table \ref{table_SOTA_vs_ours} the comparison of the accuracy metric of recent works in the SOTA for the task of kidney stone identification against our implementations.

\section{Experimental Setup}\label{Experimental_Setup}  

Three different CNN architectures are explored for our proposal: 
i) a VGG16 to examine the performance of a simple deep CNN, 
ii) a ResNet50 to consider a medium size CNN with residual connections and 
iii) a DenseNet201 for the evaluation of dense connections of a deep model to the SOTA for kidney stone identification.
These architectures were pre-trained on ImageNet, to then train them with the kidney stone dataset. 
This enables having these CNNs as baselines, and ResNet 50 and VGG16 are also used on other implementations in the current SOTA for kidney stone identification 
\cite{martinez2020towards, Towards_autoKSclass_Vincent2021, black2020deep, lopez_AssessingDL_KSclass, lopez_tiro2022boosting, VillalvazoAvila2022_ImprovedKS}.

For the prototype layer $g$, the number of PPs per class explored are $( 1, 3, 10)$ PPs per each of the six classes of the three configurations of our dataset, described in section \ref{Kidney_stone_dataset}.
The training of our implemented ProtoPNet models is a cycle consisting of three stages: ($A_1$) Updating the weights of the convolutional layers used as a feature extractor $f(x)$ per 10 epochs, using the loss function presented in \cite{ProtoPnet}.
($A_2$) Projection of the prototype layer $g$ to their corresponding closest parts after the convolutional layers $f(x)$ updates. Detailed description in supplementary material section \ref{Prototypes_projection} 
($A_3$) Update of the fully connected layer $h_1$ weights for 20 epochs (FC \& Softmax layer in Fig. \ref{fig_MI_XAI_PPs_diagram_a}). 
We repeated this cycle three times and selected the best model on the mixed test data obtained during one of the ($A_3$) phases.

Preparation for the train and test sets were 80\% (1,600 images per class) and 20\% (400 images per class), respectively. 
The patches were also ``whitened" using the mean $m_{i}$ and standard deviation $\sigma_{i}$ of the color values $I_{i}$ in each channel $(I^{w}_{i} = (I_{i}-m_{i}\sigma_{i})$, with $i=R,G,B)$. 
The different mentioned configurations for training were performed on the mixed data, described in Section \ref{Kidney_stone_dataset},
without data augmentation and with data augmentation.
The data augmentation used consists in randomly applying one data augmentation at a time to an input image, from a set of visual transformations. 
This set of transformations consists of horizontal flip, vertical flip, rotation between -180 and 180 degrees, a distortion scale of 0.4, translation up to 0.2, or symmetric padding of 50 pixels. After selecting the transformation it has a 50\% chance of been applied. 

\begin{figure*}[] 
\centering
    \includegraphics[width=1.0\linewidth]{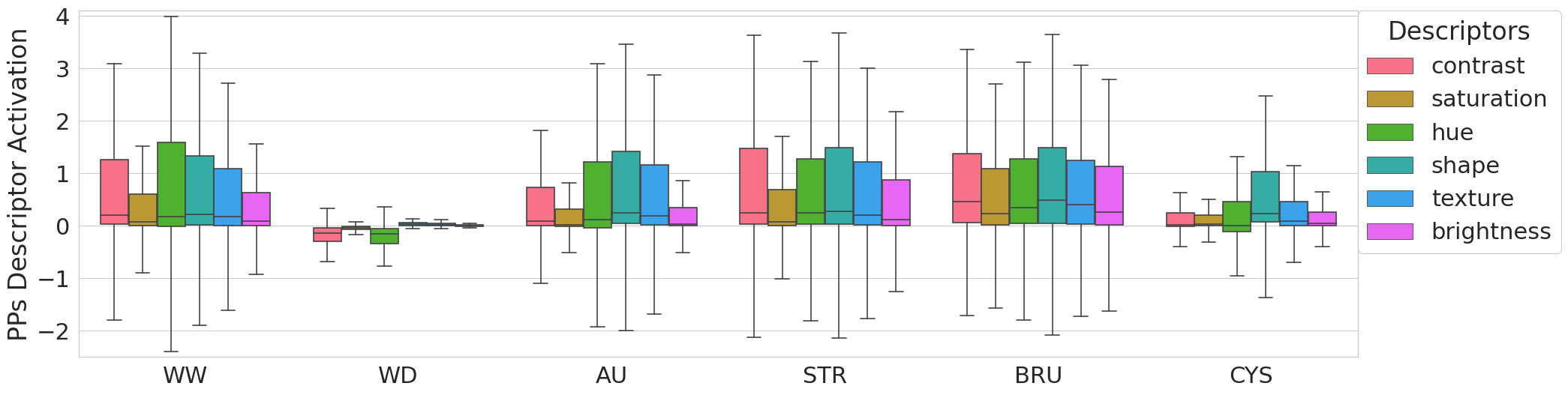} 
    \caption{\small{The six descriptor activations of the Prototypical Parts (PPs) per class, for the particular case of a ProtoPNet using DenseNet201 as the backbone and 3 Prototypical parts per class.
    }} 
\label{fig_Descriptors_per_class}
\end{figure*}

After training the different model's configurations, evaluation of the perturbations, mentioned in Section \ref{Descriptors_calculation}, is carried out as PPs \textit{descriptors} and, additionally, we evaluate the performance of the models under these perturbations (graph in the supplementary material as Fig. \ref{fig_Main_modifications}). 
To preserve clarity of results, supplemantary   \Cref{fig_Main_modifications,fig_IID_vs_OOD_general_and_per_backbone,fig_HUE_modifications}, Fig. \ref{fig_Descriptors_per_class} and \Cref{table_SOTA_vs_ours,table_IID_vs_OOD_performance}, only show the results on the mixed test data. 
%
%
Additionally, the performance of the trained models under different levels of change on the most impactful perturbation, the hue channel perturbation, is evaluated. The original change in hue explored in \cite{PP_Descriptors} applies a color jitter modification with the torchvision library 
to the input images, fixed on a change of 0.1. 
We explore a range of values for the change in the hue channel to characterize the sensitivity of the performance of our models. 
Fig. \ref{fig_HUE_modifications} illustrates the performance of one model training configuration under different hue channel settings.
%
\begin{figure*}[ht] 
    \centering
    \includegraphics[width=1.0\linewidth]{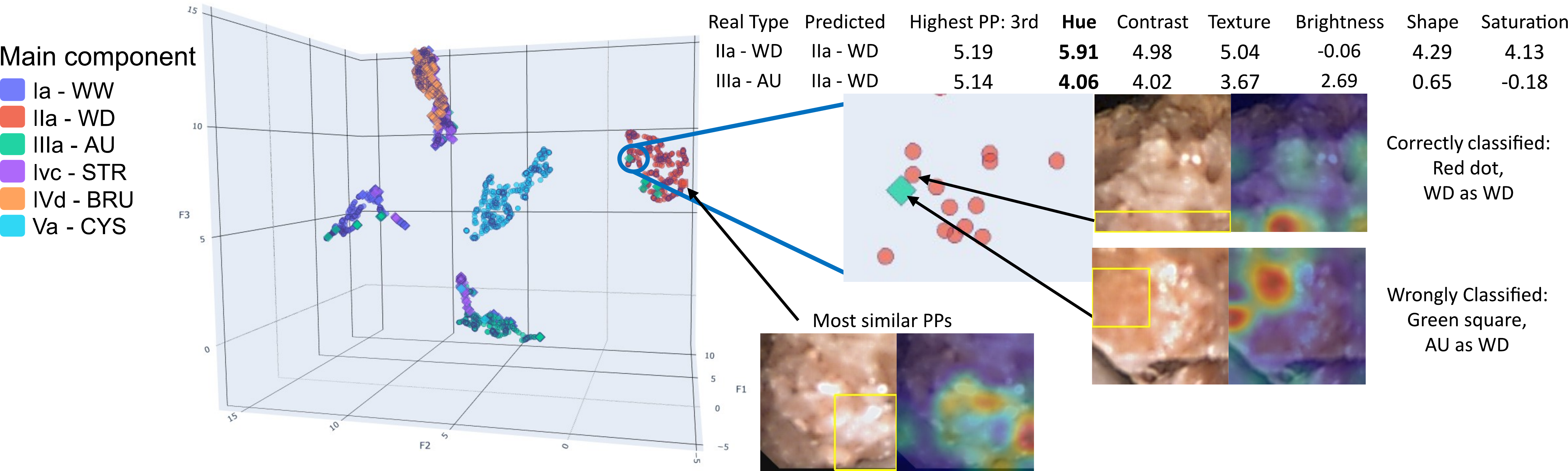}
    \caption{\small{
    UMAP of the Part-Prototypes and descriptors activations on the test patches of kidney stone images, for the particular case of a ProtoPNet using DenseNet201 as the backbone and 3 Prototypical parts per class. 
    Our approach allows obtaining separate clusters of the output classes. 
    }}
    \label{fig_umaps}
\end{figure*}
%
%

%
\begin{table}[]
\centering
\caption{
Accuracy of state-of-the-art models and three of ours for identifying kidney stones. Models trained with data augmentation.}
\label{table_SOTA_vs_ours}
\resizebox{\columnwidth}{!}{%
\begin{tabular}{ccccc}
\hline
\multicolumn{2}{c}{\textbf{Method}}            & \textbf{Surface}  & \textbf{Section} & \textbf{Mixed} \\ \hline
\multicolumn{2}{c}{Martinez, et al. \cite{martinez2020towards}} & 56.2±23.3  & 46.6±12.8           & 52.7±18.9  \\
\multicolumn{2}{c}{Estrade, et al. \cite{Towards_autoKSclass_Vincent2021}}  & 73.7±17.9  & 78.8±10.6           & 70.1±22.3  \\
\multicolumn{2}{c}{Black, et al.\cite{black2020deep}}    & 73.5±19.0  & 88.8±02.8            & 80.1±13.8  \\
\multicolumn{2}{c}{Lopez, et al.\cite{lopez_AssessingDL_KSclass}}    & 81.0±03.0   & 88.0±02.3            & 85.0±03.0   \\
\multicolumn{2}{c}{Lopez, et al.\cite{lopez_tiro2022boosting}}    & 83.2±01.2   & 90.4±04.8            & 85.6±00.1   \\
\multicolumn{2}{c}{Villalvazo-Avila, et   al.\cite{VillalvazoAvila2022_ImprovedKS}} & \textbf{88.8±02.8} & 84.4±06.0         & \textbf{96.6±00.5}        \\ \hline
\multirow{3}{*}{Ours}   & Densenet201 3pps    & 80.2±00.6        & 87.6±06.2       & 85.2±01.5               \\
           & Resnet50 1pps           & 80.9±01.3 & \textbf{91.0±03.3} & 88.2±02.1 \\
           & Vgg16 10pps             & 76.7±02.2 & 82.6±04.9          & 82.1±00.9 \\ \hline
\end{tabular}%
}
\end{table}

%
\begin{table*}[]
\footnotesize
\centering
\caption{Performance metrics of our models and convolutional neural networks used as backbones. Evaluations
are divided between tests performed with the same type of data the model was trained, considered Independent Identically
Distributed (IID) data and tests performed under perturbations, which are considered Out Of Distribution (OOD) data. 
All models were pre-trained on ImageNet. 
}
\label{table_IID_vs_OOD_performance}
\begin{tabular}{cccccccccc}
\hline
 &
   &
   &
  \multicolumn{3}{c}{IID} &
   &
  \multicolumn{3}{c}{OOD} \\
\multicolumn{3}{c}{Model} &
  Accuracy (\%) &
  Precision (\%) &
  F1 (\%) &
   &
  Accuracy (\%) &
  Precision (\%) &
  F1 (\%) \\ \cline{4-6} \cline{8-10} 
\multicolumn{3}{c}{DenseNet201} &
  \textbf{89.67±3.60} &
  \textbf{90.51±3.60} &
  \textbf{89.47±3.46} &
   &
  \textbf{58.44±28.06} &
  60.22±29.62 &
  \textbf{54.54±32.73} \\
\multicolumn{3}{c}{ResNet50} &
  83.40±5.32 &
  85.58±4.30 &
  83.11±5.21 &
   &
  45.12±20.82 &
  49.16±24.08 &
  38.25±23.47 \\
\multicolumn{3}{c}{Vgg16} &
  81.47±2.72 &
  84.71±2.64 &
  80.62±3.84 &
   &
  58.35±22.62 &
  \textbf{63.73±19.93} &
  54.38±26.30 \\
\multicolumn{3}{c}{\textbf{CNNs Average}} &
  84.84±5.29 &
  86.93±4.35 &
  84.40±5.60 &
   &
  53.97±24.62 &
  57.70±25.40 &
  49.06±28.56 \\ \hline
\multicolumn{2}{c}{\multirow{3}{*}{DenseNet201}} &
  1 PPs &
  86.29±1.91 &
  87.54±1.34 &
  86.25±1.78 &
   &
  60.31±20.32 &
  70.29±14.38 &
  57.45±23.23 \\
\multicolumn{2}{c}{} &
  3 PPs &
  85.19±1.50 &
  86.08±1.27 &
  85.21±1.42 &
   &
  59.97±19.91 &
  69.21±12.59 &
  57.66±22.14 \\
\multicolumn{2}{c}{} &
  10 PPs &
  87.29±0.92 &
  87.99±0.94 &
  87.23±0.92 &
   &
  \textbf{61.20±19.33} &
  \textbf{71.26±11.92} &
  \textbf{59.01±21.37} \\
\multicolumn{2}{c}{\multirow{3}{*}{ResNet50}} &
  1 PPs &
  \textbf{88.21±2.07} &
  \textbf{88.61±1.88} &
  \textbf{88.21±2.05} &
   &
  58.26±24.23 &
  66.48±18.53 &
  55.16±27.50 \\
\multicolumn{2}{c}{} &
  3 PPs &
  86.66±1.37 &
  87.11±1.55 &
  86.62±1.45 &
   &
  57.90±21.82 &
  63.24±19.16 &
  54.21±25.32 \\
\multicolumn{2}{c}{} &
  10 PPs &
  85.44±1.44 &
  86.16±1.15 &
  85.43±1.42 &
   &
  56.31±22.66 &
  64.81±16.55 &
  53.77±25.52 \\
\multicolumn{2}{c}{\multirow{3}{*}{Vgg16}} &
  1 PPs &
  81.78±1.60 &
  83.38±1.94 &
  81.82±1.62 &
   &
  51.78±19.02 &
  53.82±19.40 &
  46.63±22.77 \\
\multicolumn{2}{c}{} &
  3 PPs &
  82.21±3.33 &
  82.92±3.82 &
  81.84±3.75 &
   &
  51.23±19.40 &
  54.09±19.80 &
  45.88±23.20 \\
\multicolumn{2}{c}{} &
  10 PPs &
  82.08±0.90 &
  83.39±0.92 &
  82.13±0.87 &
   &
  54.69±18.05 &
  56.48±17.56 &
  49.89±21.51 \\
\multicolumn{3}{c}{\textbf{ProtoPNets Average}} &
  85.02±2.83 &
  85.91±2.66 &
  84.97±2.89 &
   &
  56.85±20.63 &
  63.30±17.90 &
  53.30±23.83 \\ \hline
\end{tabular}%
\end{table*}
\section{Results and Discussion}\label{Results}

\subsection{Implementation details} \label{Implementation_Details}
 Experiments were carried out to explore different backbones for the feature extractor network $f(x)$, the numbers of PPs per class and the significance of data augmentation. This, is to evaluate the model's performance and the obtained explanations. 
%
In contrast to CNNs, which are black-box classifiers, our proposal provides explanations for input images in three different ways: \textbf{i)} by showing the activation area, where each PP is identified. \textbf{ii)} The corresponding representative example image to each activated PPs. \textbf{iii)} Quantitative evaluations of how relevant six different visual descriptors are for the activation of each PP.
\subsection{Visualization of a case example of explanations} \label{Explanation_visualization}
%
%
In Fig. \ref{fig_umaps}, an example of our model classifications is shown. 
We found that averaging the global scores of the PPs descriptors per class results in a hierarchy of descriptor significance, as depicted in Fig. \ref{fig_Descriptors_per_class}. 
The WD class reveals a unique pattern of descriptor activation.
An examination of several instances shows that the average activation of a class descriptors can hint at misclassifications.
As shown in Fig. \ref{fig_umaps}, a green diamond represents a misclassified AU image, which is misidentified as WD due to its activations corresponding to those of the WD class. This causes the image to appear within the WD class cluster on the UMAP plot.
We use a UMAP visualization to present PPs activations for input images in the context of the top three discriminative dimensions. 
This UMAP highlights class separability for each output class of the ProtoPNet, as showcased in Fig. \ref{fig_umaps}.
%
%
The use of descriptors mitigates the cases for visually similar PPs by providing details on the characteristics most relevant for each PP \cite{PP_Descriptors}. 
%
Additionally, learning PPs from patches of kidney stone images allows observing PPs are learned proportionally to the architecture of the CNN used. 
\subsection{Performance results} \label{Performance_results}
A summary of the accuracy of the different configurations of models trained is displayed in Table \ref{table_IID_vs_OOD_performance}. Tests carried out on the original mixed test dataset, corresponding to the training mixed dataset, are considered Independent Identically Distributed (IID) data, tests performed under perturbations applied to the original test dataset are considered Out Of Distribution (OOD) data and named accordingly in Table \ref{table_IID_vs_OOD_performance}.
Usage of data augmentation yielded an average 3\% improvement accuracy compared to the non-augmented approach consistently across different evaluations. To ease presentation, we report results just on models trained with data augmentation.

The performance of our implemented ProtoPNet models is comparable with its corresponding non-interpretable CNN backbone models as also appreciated in Table \ref{table_IID_vs_OOD_performance}. 
The average accuracy of our implementations keeps competitiveness with a difference not bigger than $\le0.18\%$, when compared with the average accuracy of the baseline CNN models.
The mean values and standard deviations displayed in Table \ref{table_IID_vs_OOD_performance} are computed based on the configurations of the models outlined therein.
It is worth noticing that Table \ref{table_IID_vs_OOD_performance} shows our interpretable models to outperform traditional CNNs from the SOTA on kidney stone classification when evaluated on perturbed input images. 
Specifically, our models achieve 1.9\% higher accuracy, 5.7\% higher precision, and 3.2\% higher F1 scores compared to traditional models. 

\section{Conclusions and Future Directions}\label{Conclusions}
We showed that by adapting and fine tunning CNN models into ProtoPNets 
is possible to convert considered black-box CNN models into interpretable ones, bringing detailed and faithful explanations.
%
The main visual features in the input image that produce the activation of PPs in particular and in the average cases per class and per dataset are able to be extracted from the evaluated behavior of the network, as shown in Fig. \ref{fig_Descriptors_per_class}.  
This case-based reasoning approach allows to generate explanations and additional activation details enabling urologists to use these models as assistance tools for the MCA.
More importantly, we showed evidence for how to use the overall behavior of descriptors per class in combination with UMAP projection of the behavior of the model to identify the reasons behind misclassification, as shown in Fig. \ref{fig_umaps}.
Also, the UMAP visualization illustrates the blending of certain classes that share similar color and texture characteristics. 
To enhance cluster separability a metric learning approach could be explored. This to penalizes close PPs from different classes, encourages proximity for PPs of the same class but avoids their collapse.
%


Upon examining Table 3, on one hand, it is evident that despite a minor decrease in average performance for IID evaluations in PPs models compared to their corresponding CNNs. This could be due to restricting the classification to a few PPs per class, in comparison CNN models use a few hundred features. 
On the other hand, PPs models consistently exhibit reduced variance in performance metrics. Additionally, these models consistently perform better in OOD evaluations, suggesting that PPs models tend to be more robust than their CNN counterparts. This characteristic could be attributed to a regularization effect resulting from measuring the similarity between the learned PPs and the image features.
PPs models inherit most of the performance characteristics of CNNs, indicating that methods to enhance CNN performance should also improve PPs models, thus expanding potential improvements. 

In future research, the model's reasoning and reporting mechanisms could be refined to align more closely with the structured lexicon employed by urologists.
This could be achieved by incorporating natural language descriptions for the PPs identified and their \textit{descriptors} to make explanations more accessible to specialists and a broader range of users.

\section*{Compliance with Ethical Standards}
The images were captured in medical procedures following
the ethical principles outlined in the Helsinki Declaration of
1975, as revised in 2000, with the consent of the patients.
 
\section*{Acknowledgments}
The authors extend their gratitude to the AI Hub and the Centro de Innovación de Internet de las Cosas at Tecnológico de Monterrey for generously providing access to an NVIDIA DGX computer, which facilitated the experiments conducted in this paper. Additionally, we would like to express our gratitude to CONACyT for the Ph.D. scholarships for Daniel Flores-Araiza and Francisco López-Tiro.
%
{\small
\bibliographystyle{ieee_fullname}
\bibliography{ISBI2023}
}
%
\clearpage
\setcounter{section}{0} 
\setcounter{figure}{0} 
\renewcommand{\thefigure}{S\arabic{figure}} 
\renewcommand{\thesection}{S\arabic{section}} 



\section{Supplementary Material }\label{Supplementary}
\subsection{Performance under visual perturbations} \label{visual_perturbations_performance}
Accuracy evaluated on the mixed test dataset, for the different
ProtoPNet backbones and the number of PPs learned was obtained to identify the level of loss in performance of the models learning PPs. 
With this evidence for the kidney stone classification task made available, future users will have a reference for which visual characteristics PPs models tend to be more susceptible to and to which degree. 
It is found that hue perturbation on the input images is on average the
perturbation with the most dramatic loss in performance for models learning PPs training with the Kidney Stones datasets. In contrast, this type of architecture tends to be robust against moderated changes in the brightness and saturation of the input images. 
With this evidence, a call for precaution when dealing with assumed small perturbation on the input images \cite{PP_Descriptors} is implemented.
%

\begin{figure*}[ht] 
    \centering
    \subfloat[\small{CNNs}]{
    \label{fig_a_Backbones_IID_vs_OOD}
    \includegraphics[width=0.20\linewidth]{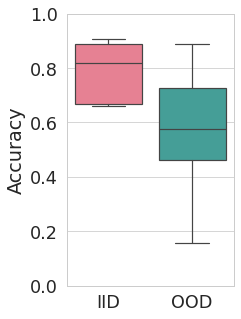} 
    } 
    \subfloat[\small{ProtoPNets}]{
    \label{fig_b_ProtoPNets_IID_vs_OOD}
    \includegraphics[width=0.20\linewidth]{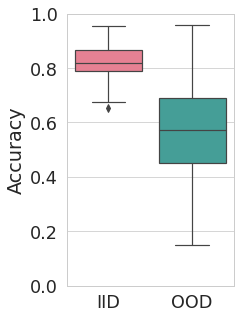} 
    } 
    \subfloat[\small{ProtoPNets per CNN backbone}]{
    \label{fig_c_IID_vs_OOD_per_backbone}
    \includegraphics[width=0.49\linewidth]{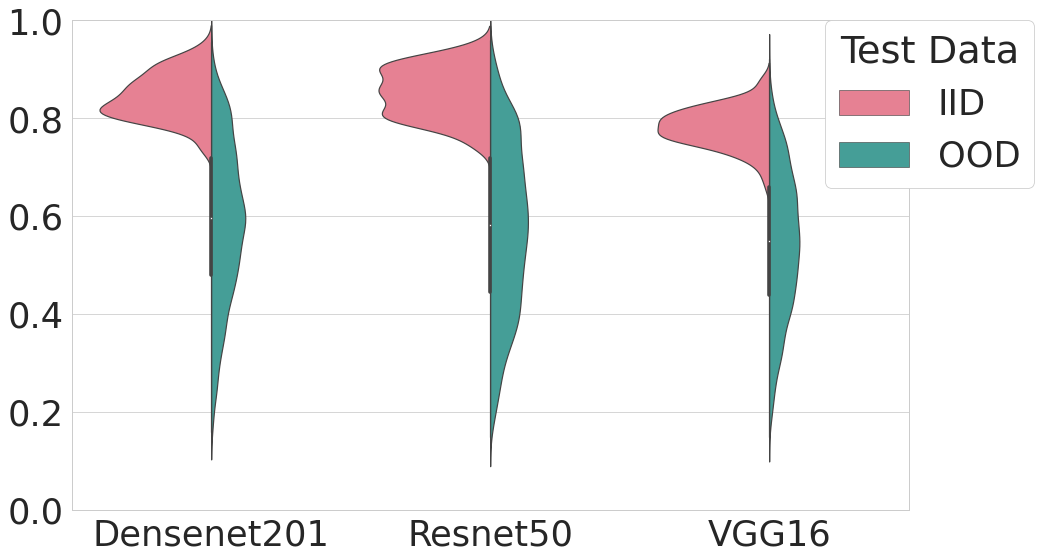}
    }
    \caption{\small{ 
    (a) Accuracy of CNN models evaluated separated on Independent and Identically Distributed (IID) test data and Out Of Distribution (OOD) tests.
    (b) Accuracy of ProtoPNet models evaluated on IDD and OOD test data.
    (c) Same as in "b" for each of the three backbones used. 
    }}
    \label{fig_IID_vs_OOD_general_and_per_backbone}
\end{figure*}

Also, the performance of the traditional CNN used as feature extractors were explored under the same perturbations, these results are shown in the first four rows under the OOD performance metrics in Table \ref{table_IID_vs_OOD_performance} for each of the CNN backbones explored and their performance for IID test data and OOD is summarized in Fig. \ref{fig_a_Backbones_IID_vs_OOD}. These same evaluations are shown for the different backbones of PPs models trained in Fig. \ref{fig_b_ProtoPNets_IID_vs_OOD}. It is observed PPs models present a lower standard deviation when compared to their CNN counterparts for IID evaluations, per this behavior for each of the different backbones used by the models, as seen in Fig. \ref{fig_c_IID_vs_OOD_per_backbone}. 

\begin{figure*}[hbtp]
\centering
    \includegraphics[width=0.90\linewidth]{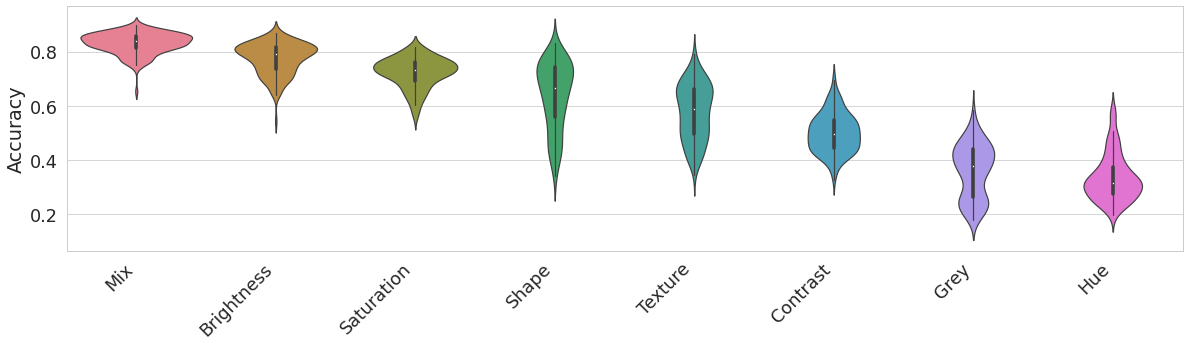} 
    \caption{\small{Accuracy evaluated on the test dataset, averaging the different ProtoPNet backbones and the number of Prototypical parts learned. It is found that hue perturbation on the input images is on average the perturbation with the most dramatic loss in performance for ProtoPNet models training with the Kidney Stones dataset. In contrast, the ProtoPNet architecture tends to be robust against moderated changes in the brightness and saturation of the input images.
    }} 
\label{fig_Main_modifications}
\end{figure*}

\begin{figure*}[h] 
\centering
    \includegraphics[width=0.90\linewidth]{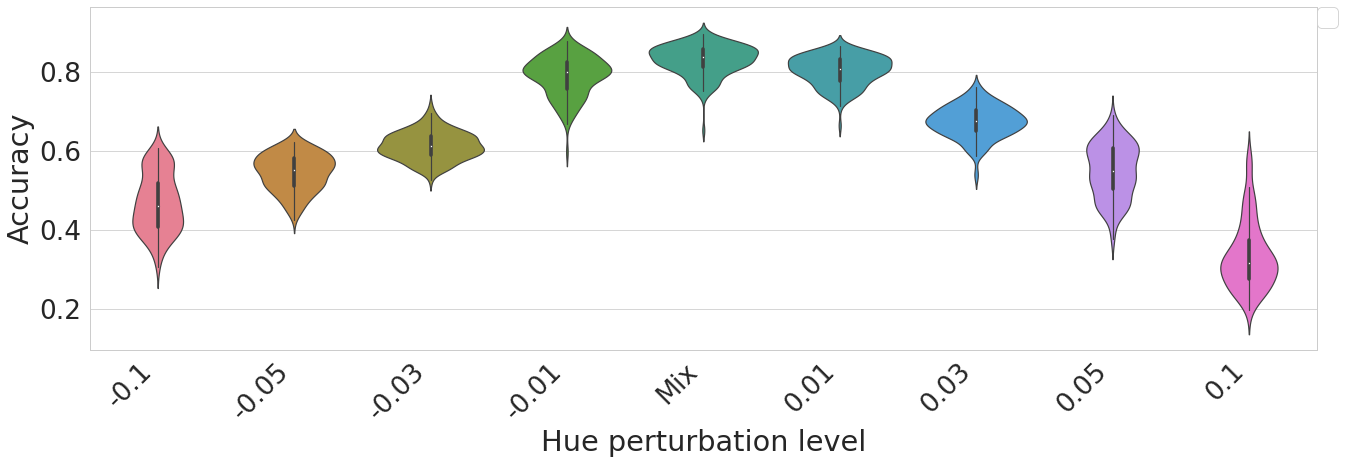}
    \caption{\small{Accuracy evaluated on mixed test dataset under modifications of the images hue channel, with values between -0.1 and 0.1. These evaluations reflect the level of robustness ProtoPNet achieves with a DenseNet201 CNN backbone with 3 part-prototypes per class, under the perturbation that most affect the model.}} 
\label{fig_HUE_modifications}
\end{figure*}

%


\subsection{Prototypes projection details} \label{Prototypes_projection}

The Projection of Prototypical parts (PPs) is the intermediate step $A_2$, mentioned in Section \ref{Experimental_Setup}, performed in the training process that allows visualization of the learned PPs. 
In this step, each prototype $p_j$ is assigned the value of its nearest latent training patch $f(x)_l$ 
from all the images of the same class $k$ initially assigned to $p_j$.
Therefore, this distance is $d_{j,l,k} = \lVert  p_{j,k} - f(x)_{l,k} \rVert_{2}$. 
In this way, each abstract PPs learned conceptually equates to one training image patch.
Allowing the faithful visualization of the learned PPs, use to generate the visual explanations as well as the final output classification given by the model.
Mathematically, for the projection prototype $p_j$ of class $k$, i.e., $p_j \in P_k$, we perform the following update:

\begin{equation}\label{Global_descriptors}
    p_{j,k} \xleftarrow{} \operatorname*{arg\,min}_{ f(x)_{l,k} \in S_{train} } \lVert  p_{j,k} - f(x)_{l,k} \rVert_{2}  
\end{equation}

where $ S_{train} = \{ f(x)_{l,k} : f(x)_{l,k} \in f(x_k)$ for  all training images $S_{train}$,  $(x, y) : y \in k \}$.

\end{document}